\documentclass[letterpaper]{article} 
\usepackage{aaai2026}  
\usepackage{times}  
\usepackage{helvet}  
\usepackage{courier}  
\usepackage[hyphens]{url}  
\usepackage{graphicx} 
\usepackage{natbib}  
\usepackage{caption} 
\usepackage{algorithm}
\usepackage{algorithmic}

\usepackage{newfloat}
\usepackage{listings}
\DeclareCaptionStyle{ruled}{labelfont=normalfont,labelsep=colon,strut=off} 
\floatstyle{ruled}
\newfloat{listing}{tb}{lst}{}
\floatname{listing}{Listing}
\usepackage{booktabs,multirow,makecell,xcolor}
\usepackage{amsfonts,amsmath}
\usepackage{enumitem}

\title{Generalizable and Efficient Automated Scoring with a Knowledge-Distilled Multi-Task Mixture-of-Experts}
\author {
    Luyang Fang\textsuperscript{\rm 1,\rm 2},
    Tao Wang\textsuperscript{\rm 1},
    Ping Ma\textsuperscript{\rm 1},
    Xiaoming Zhai\textsuperscript{\rm 2}\thanks{This is the corresponding author.}
}
\affiliations {
    \textsuperscript{\rm 1}Department of Statistics, University of Georgia\\
    \textsuperscript{\rm 2}AI4STEM Education Center, University of Georgia\\
    Luyang.Fang@uga.edu, Tao.Wang2@uga.edu, pingma@uga.edu, Xiaoming.Zhai@uga.edu
}

\usepackage{bibentry}

\begin{document}

\maketitle

\begin{abstract}
Automated scoring of written constructed responses typically relies on separate models per task, straining computational resources, storage, and maintenance in real-world education settings. We propose UniMoE-Guided, a knowledge-distilled multi-task Mixture-of-Experts (MoE) approach that transfers expertise from multiple task-specific large models (teachers) into a single compact, deployable model (student). The student combines (i) a shared encoder for cross-task representations, (ii) a gated MoE block that balances shared and task-specific processing, and (iii) lightweight task heads. Trained with both ground-truth labels and teacher guidance, the student matches strong task-specific models while being far more efficient to train, store, and deploy. Beyond efficiency, the MoE layer improves transfer and generalization: experts develop reusable skills that boost cross-task performance and enable rapid adaptation to new tasks with minimal additions and tuning.
On nine NGSS-aligned science-reasoning tasks (seven for training/evaluation and two held out for adaptation), UniMoE-Guided attains performance comparable to per-task models while using $\sim$6$\times$ less storage than maintaining separate students, and $87\times$ less than the 20B-parameter teacher.
The method offers a practical path toward scalable, reliable, and resource-efficient automated scoring for classroom and large-scale assessment systems.
\end{abstract}

\begin{links}
    \link{Code}{https://github.com/LuyangFang/UniMoE}
\end{links}

\section{Introduction}\label{sec:intro}

Automated scoring systems have become indispensable in modern educational assessment by enabling efficient evaluation of students’ open-ended responses, particularly in science and STEM domains \cite{page1966imminence,dikli2006overview,moharreri2014evograder,jescovitch2021comparison}. As curriculum frameworks like the Next Generation Science Standards (NGSS) promote complex performance tasks to assess multidimensional understanding \cite{Harris2024Creating}, demand for reliable, scalable automated scoring grows.
Transformer-based deep learning models, including large language models (LLMs), deliver strong accuracy for automated essay scoring (AES) and short-answer grading (ASAG), enhancing measurement fidelity and operational efficiency \cite{zhai2022applying,lee2024applying,latif2024knowledge}. However, the dominant cost driver remains the maintenance and deployment of numerous per-item or per-task models. In classroom assessments, it is standard practice to train models per task to ensure alignment with item-specific rubrics and expert criteria. This proliferation of models generates many artifacts requiring storage, monitoring, and redeployment; the burden is especially severe in classroom assessment systems and statewide or district-scale testing programs where storage constraints and inference speed are critical \cite{liu2021towards,zhai2022applying,reidy2023efficient}.

Methods like parameter-efficient tuning \citep{han2024parameter,mahmoud2024automatic} reduce the number of trainable parameters and update size. For example, LoRA often matches full fine-tuning without added inference latency by inserting low-rank matrices while freezing the backbone \citep{hu2021lora}. However, these approaches still require loading and serving the full base model at inference, so memory footprint, cold-start time, and per-task model variants remain the dominant cost drivers - especially when many tasks must be supported concurrently (a common reality for educational platforms).
We therefore pose the central question: for challenging cross-prompt trait-scoring scenarios, including those in AES/ASAG \cite{yadav2023ties,katuka2024investigating}, \textit{can a single and compact model support multiple scoring tasks without material loss in performance?}

To answer this question, we propose \textsc{UniMoE}: a knowledge-distilled multi-task method based on a Mixture-of-Experts (MoE) architecture that trains a single compact model for multiple tasks. A multi-task \citep{zhang2018overview,fang2025efficient} backbone learns representations common to automated scoring, while lightweight task heads produce rubric-specific predictions (e.g., holistic and trait scores). Between them, a gated MoE module \citep{masoudnia2014mixture,fedus2022review} routes each example to experts with different weights, preserving item/trait idiosyncrasies and mitigating negative transfer across heterogeneous tasks. 
Practically, \textsc{UniMoE} consolidates storage and serving into a single backbone, minimizes per-task additions, and adapts to new tasks more reliably than standard multi-task models. By providing specialized capacity for rubric-specific patterns, the MoE layer reduces inter-task interference and enhances transfer learning: new tasks reuse shared representations while leveraging specialized experts, requiring only small task-specific components without destabilizing prior tasks. The result is faster, lower-cost adaptation at scale without sacrificing rubric fidelity.

In settings where high-capacity per-task reference models (`teachers') exist or can be trained offline, but deployment requires a single lightweight model, we extend \textsc{UniMoE} to \textsc{UniMoE-Guided} via knowledge distillation \citep{gou2021knowledge,fang2025knowledge}. The compact multi-task model used at inference serves as the `student'. For each task, the teacher provides soft guidance (e.g., probability predictions), and the student minimizes a convex combination of the supervised loss and a distillation loss that aligns its outputs with the teacher's; the distillation term acts as a regularizer. This transfers knowledge from the teachers into the student, narrowing the gap to task-specialized models while preserving a single low-latency model with substantially lower storage and serving costs.

We evaluate \textsc{UniMoE-Guided} on NGSS-aligned, multi-label science-reasoning tasks from the PASTA corpus, using seven primary tasks for model comparison and held-out tasks to probe generalization. Briefly, results show that \textsc{UniMoE-Guided} achieves comparable results to training separate models, while being $\sim$6$\times$ smaller than maintaining separate models and $87\times$ smaller than the teacher model. It also remains deployment-friendly, underscoring its practical deployability. Additionally, the generalization tasks show that \textsc{UniMoE-Guided} achieves significantly better performance than simple MTL, emphasizing its superior generalization capability in handling unseen tasks.

The proposed \textsc{UniMoE-Guided} method supports resource-efficient AI for educational automated scoring and other cost-sensitive domains, contributing to advancements in scalable, efficient model deployment. Our key contributions are summarized below:
\begin{itemize}[itemsep=1pt, topsep=2pt]
    \item \textbf{Single deployable multi-task scorer.} A shared encoder plus a task-aware MoE and lightweight heads balance shared structure with rubric-specific specialization, mitigating negative transfer across diverse assessments.
    \item \textbf{Teacher-guided compression.} Distillation from task-specialized teachers into one compact student yields the highest average reliability on our seven-task benchmark, while reducing storage by $\sim$6$\times$ versus maintaining separate per-task students and by $87\times$ versus the 20B teacher.
    \item \textbf{Robust extension to new tasks.} Adding a new head and lightly tuning a small portion of the MoE integrates new assessments without retraining the backbone, outperforming plain MTL and narrowing the gap to per-task models on held-out tasks.
\end{itemize}

\section{Related Work}\label{sec:relate}
\noindent\textbf{Multi-Task Learning (MTL).}
MTL has long been explored as a means to improve generalization by leveraging inductive biases shared across related tasks \cite{caruana1997mtl}. Traditional approaches rely on hard or soft parameter sharing within a shared encoder, while more recent methods incorporate optimization-aware strategies, adaptive task weighting, and explicit modeling of inter-task relationships, particularly in the era of large foundation models \cite{yu2024mtlsurvey}. In the context of educational assessment, jointly training across rubric dimensions (multi-label) and prompts (multi-task) can increase efficiency but risks negative transfer when task distributions diverge. To mitigate this, modern MTL designs often combine a shared backbone with lightweight task-specific modules or routing mechanisms to preserve both efficiency and task fidelity \cite{jacob2023onlinekd,auty2024crosskd}. This motivates our choice of a shared encoder augmented with task-aware specialization capacity.

\noindent\textbf{Mixture-of-Experts (MoE) Architectures.}
MoE models address the limitations of fixed shared capacity in MTL by increasing total model capacity while keeping per-example computation low through conditional routing of tokens to a small subset of experts \cite{shazeer2017moe}. Large-scale deployments such as GShard and Switch Transformers demonstrated the scalability of gating paired with auxiliary load-balancing losses \cite{fedus2022switch}. Subsequent advances, including expert-choice routing \cite{zhou2022expertchoice} and state-of-the-art MoE implementations in Mixtral, DeepSeek-V2/V3, and Qwen-2.x \cite{mixtral2024,deepseekv2,deepseekv3,qwen2report,qwen25report}, highlight the potential of weighted activation to enable diverse specializations without prohibitive costs. 

\noindent\textbf{Knowledge Distillation (KD).}
Even with efficient architectures like MoE, deploying large models in educational settings often requires further compression. KD offers a principled solution by transferring the behavior of high-capacity teacher models into smaller, faster student models \cite{gou2021knowledge,fang2025knowledge,fang2024bayesian}. The field has evolved from early logit-matching to feature-based transfer, self-distillation, and task-aware distillation strategies. For multi-label scenarios such as rubric scoring, specialized KD methods mitigate label competition by distilling one-vs-rest probabilities or label-wise embeddings \cite{yang2023mlkd}, while in multi-task contexts, cross-task KD improves stability and transferability under distribution shifts \cite{jacob2023onlinekd,auty2024crosskd}. In educational assessment, recent studies have applied KD to fine-tune LLMs for scoring under latency and fairness constraints \cite{latif2024scoring,misgna2024aes}. Building on this foundation, our approach distills knowledge from multiple strong task-specific teachers into a single multi-task MoE student.

\begin{figure*}[ht]
    \centering
    \includegraphics[width=0.99\linewidth]{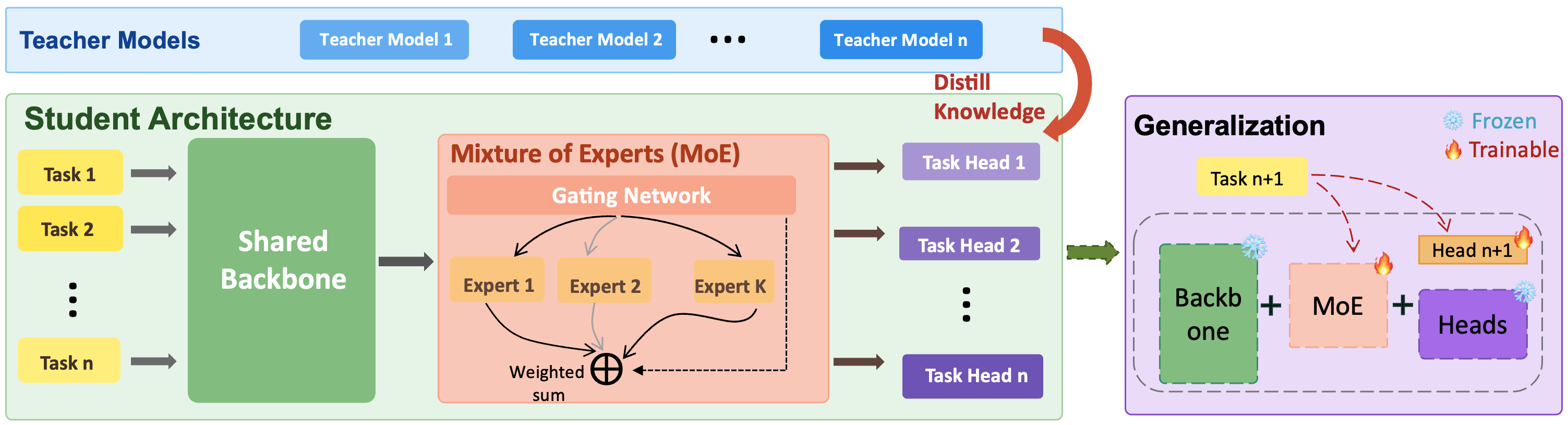}
    \caption{Workflow of \textsc{UniMoE} and its teacher-guided variant \textsc{UniMoE-Guided}. Left and center: multiple tasks share a single backbone; a MoE module with a learned gating network routes representations to a small subset of experts and aggregates them by a weighted sum; the outputs feed lightweight task heads. In \textsc{UniMoE-Guided}, teacher models provide soft targets during training. Right: for a new task, the backbone and existing heads stay frozen, and only a new head and optionally a small portion of the MoE are trained, enabling rapid, low-overhead adaptation.}
    \label{fig:workflow}
\end{figure*}

\section{Methodology}\label{sec:method}

In this section, we describe the proposed \textsc{UniMoE-Guided} method, which constructs a single student model for automated scoring across multiple educational tasks. The resulting model consists of three components: a shared encoder that captures general linguistic representations, a task-aware Mixture-of-Experts (MoE) layer that activates a small set of specialized experts for each response, and task-specific heads that generate rubric-aligned predictions. To train the student, \textsc{UniMoE-Guided} combines authentic labels with KD from high-capacity teacher models, allowing the student to approach teacher-level accuracy while remaining compact and practical for classroom deployment.

\subsection{MoE-Based Multi-Task Learning with Knowledge Distillation}

We model automated scoring of middle-school science explanations as multi-label prediction across $T$ NGSS-aligned tasks. For a student response $X_t$ from task $t$ with binary rubric indicators \(y_t \in \{0,1\}^{D_t}\) the model outputs logits \(z_t(X_t) \in \mathbb{R}^{D_t}\) and per-indicator probabilities
\[
p_t(X_t) = \sigma\!\left(z_t(X_t)\right),
\]
where \(\sigma(\cdot)\) denotes the element-wise sigmoid function.

\noindent\textbf{Architecture.}
The proposed architecture, \textsc{UniMoE-Guided}, includes (i) a shared Transformer encoder, (ii) a MoE block with task-aware routing, and (iii) compact task-specific heads. The shared encoder $f_\theta(\cdot)$ shares parameters across all tasks to provide a common linguistic representation, enabling a single deployable checkpoint. The MoE block combines multiple experts via a gating network that produces varying weights per input, allowing experts to specialize in rubric-specific language and reasoning patterns.
Lastly, the task-specific heads are tailored to each task, maintaining rubric fidelity and strong per-task accuracy. The heads remain small, so adding a new rubric increases storage only by a lightweight head rather than a full model.
We introduce the details of the three parts below.

\noindent\textbf{Shared encoder.} We use a pre-trained BERT-base model \citep{devlin2019bert} as a shared text encoder, fine-tuned on our tasks, that maps a student response \(X\) to contextual representations \(H_t = f_{\theta}(X_t)\), where \(H_t \in \mathbb{R}^{L \times d}\) with sequence length \(L\), and hidden size \(d\). This task-agnostic encoder captures linguistic patterns common across rubrics and is reused for every task, reducing storage and simplifying deployment.

\noindent\textbf{MoE with task-aware routing.}
We append a MoE block after the encoder to enable dynamic routing of representations to specialized experts, improving task-specific adaptation. 
The MoE block comprises \(M\) experts \(\{E_j\}_{j=1}^M\), where each expert is a two-layer feed-forward network with nonlinearity and dropout, applied token-wise to the contextual representation \(H_t\), resulting $E_j(H_t) \in \mathbb{R}^{L \times d}$.

A task-aware gating network produces mixture weights over experts. 
We form the gate input by concatenating a pooled sequence summary $h_{\text {t,pool }}=\pi(H_t) \in \mathbb{R}^d$ with a learned task embedding $e_t$, yielding $m_t=\left[h_{\text {t,pool }} ; e_t\right]$. The gating network $g(\cdot)$ maps $m_t$ to logits $G_t= \left(g_{t,1}, \ldots, g_{t,M}\right)^{\top} \in \mathbb{R}^M$, which are converted to weights $\alpha_t=\left(\alpha_{t,1}, \ldots, \alpha_{t,M}\right)^{\top}$ by a softmax,
$$
\alpha_{t,j}=\frac{\exp (g_{t,j})}{\sum_{j=1}^M \exp (g_{t,j})}.
$$
These weights are shared across tokens, yielding MoE-enhanced representation \(\tilde{H}_t\):
$$
\tilde{H}_t \;=\; \sum_{j=1}^M \alpha_{t,j}\, E_j(H_t).
$$

To discourage expert collapse and promote coverage, we add a load-balancing penalty that encourages the average gate weights to match the uniform distribution.
For the responses of size $B$, let \(\alpha_{j}^{(b)}\) be the weight assigned to expert $j$ for example $b$, and define 
$\bar{\alpha}_{j}=\frac{1}{B} \sum_{b=1}^B \alpha_{j}^{(b)}$. The load-balancing penalty is 
$$
\mathcal{L}_{\mathrm{LB}} \;=\; \frac{1}{M}\sum_{j=1}^{M}\!\left(\bar{\alpha}_{j}-\frac{1}{M}\right)^{2}.
$$
This regularizer encourages broader expert utilization. The mechanisms of expert structure, task-aware gating, and explicit balancing are essential for stable specialization across heterogeneous rubrics.

\noindent\textbf{Task-specific heads.} For each task $t$ with $D_t$ rubric labels, a small MLP head $h_t(\cdot)$ maps the MoE-enhanced representation $\tilde{H}$ to logits $z_t \in \mathbb{R}^{D_t}$.
Only heads are task-specific, so adding a rubric requires instantiating a new head (and, if desired, modest MoE tuning) rather than a new model.

\noindent\textbf{Training objectives.} 
The NGSS-aligned assessments are multi-label: each label is binary, and multiple labels may be present simultaneously. We therefore optimize a binary cross-entropy (BCE) objective:
\begin{align*}
    \mathcal{L}_{\text{task}} = E_{(t,X_t)}( \sum_{c=1}^{D_t} \big[&-\, y_{t,c}\,\log \sigma(z_{t,c}) \\
    &- (1 - y_{t,c})\,\log(1 - \sigma(z_{t,c}))\big] ),
\end{align*}
where \(y_t \in \{0,1\}^{D_t}\) are the ground-truth rubric indicators for task $t$. In practice, we implement the expectation as an average over tasks and samples. This loss directly models the presence or absence of multiple skills or evidence types in a single response.

\noindent\textbf{Knowledge distillation.}
To reduce storage while preserving agreement with expert scoring, we integrate KD from pre-trained teacher models. Given teacher logits $z^{\operatorname{(teacher)}}$, which are precomputed and loaded during training, and a temperature $\tau>0$, we define softened targets $q^{\tau}=\sigma\left(z^{\operatorname{(teacher)}} / \tau\right)$ and softened student outputs $p^{(\tau)}=\sigma\left(z_t / \tau\right)$, and minimize a distillation BCE:
$$
\mathcal{L}_{\mathrm{KD}}= E_{(t,X_t)}\left( \operatorname{BCE}\left(p^{(\tau)}, q^{\tau} \right) \cdot \tau^2 \right).
$$

KD transfers rubric-specific decision boundaries to the student, stabilizes learning when some labels are sparse, and enables efficient deployment without reintroducing teachers at inference. 

\noindent\textbf{Overall training objective.}
We minimize the average per-example objective across all tasks and samples, plus the load-balancing penalty, and the distillation regularization:
$$
\mathcal{L}=\mathcal{L}_{\text {task }}+\lambda_{\mathrm{LB}} \mathcal{L}_{\mathrm{LB}}+\lambda_{\mathrm{KD}} \mathcal{L}_{\mathrm{KD}} ,
$$
where $\lambda_{\mathrm{LB}}$ and $\lambda_{\mathrm{KD}}$ are weights controlling the contribution of load-balancing penalty and distillation regularization.


\subsection{Generalization}

For a new assessment task, we extend the trained multi-task MoE model in two steps that preserve prior competence while enabling fast adaptation.
We append a lightweight, task-specific prediction head for the new task. During adaptation, the shared backbone remains frozen to preserve linguistic representations; we update only the MoE layers (experts and router) and the newly added head. Training uses labeled data from the new task together with a small rehearsal subset from previously seen tasks to mitigate forgetting. When teacher signals are available, we reuse the same distillation objective defined in base training; otherwise, we optimize the supervised task loss with the existing load-balancing regularizer to prevent expert collapse. Only a small fraction of parameters is updated, enabling rapid, stable inclusion of new assessments while maintaining performance on earlier tasks.

In summary, \textsc{UniMoE-Guided} combines a shared encoder that provides transferable linguistic representations, a task-aware, load-balanced MoE that delivers scalable specialization, and compact task heads that capture rubric differences with minimal storage. KD then consolidates teacher knowledge into a single deployable student, well-suited to resource- and privacy-constrained classroom settings.

\section{Dataset Details}\label{sec:dataset}

We use pre-existing responses from U.S. middle-school classrooms (grades 6–8) to nine NGSS-aligned assessment tasks developed under the NGSA initiative \cite{Harris2024Creating,PASTA2023}. Each response is scored across multiple rubric dimensions, with each label treated as binary, indicating whether a given dimension is satisfied. The tasks were designed to elicit evidence of three-dimensional learning as defined by the NGSS, integrating Disciplinary Core Ideas (DCIs), Crosscutting Concepts (CCCs), and Science and Engineering Practices (SEPs). In particular, they align with MS-PS1-2, where students analyze and interpret data on the properties of substances before and after interaction to determine whether a chemical reaction has occurred \cite{national2013next}.

\begin{table}[t]
\centering
\begin{tabular}{l|c|c|c}
\toprule
\textbf{Dataset} & \textbf{No. Labels} & \textbf{Training size} & \textbf{Testing size} \\
\midrule
Task 1 & 4 & 955 & 239 \\
Task 2 & 4 & 666 & 167 \\
Task 3 & 3 & 958 & 240 \\
Task 4 & 10 & 956 & 240 \\
Task 5 & 5 & 836 & 210 \\
Task 6 & 6 & 653 & 164 \\
Task 7 & 5 & 956 & 239 \\
Task 8 & 5 & 956 & 240 \\
Task 9 & 3 & 1111 & 278 \\
\bottomrule
\end{tabular}%
\caption{Statistics of the nine NGSS-aligned assessment tasks from the PASTA corpus.}
\label{table:data}
\end{table}

For each task, approximately 1,200 students' responses were randomly selected from a large pool of responses, and after data cleaning, each task included slightly fewer responses; exact per-task counts and train/test splits are reported in Table~\ref{table:data}  \cite{zhai2022applying}. All responses were anonymized, and no demographic variables were included. The sample reflects geographic diversity across schools to be statistically representative of the broader U.S. population.

\begin{figure}[hbt!]
    \centering
    \includegraphics[width=0.99\linewidth]{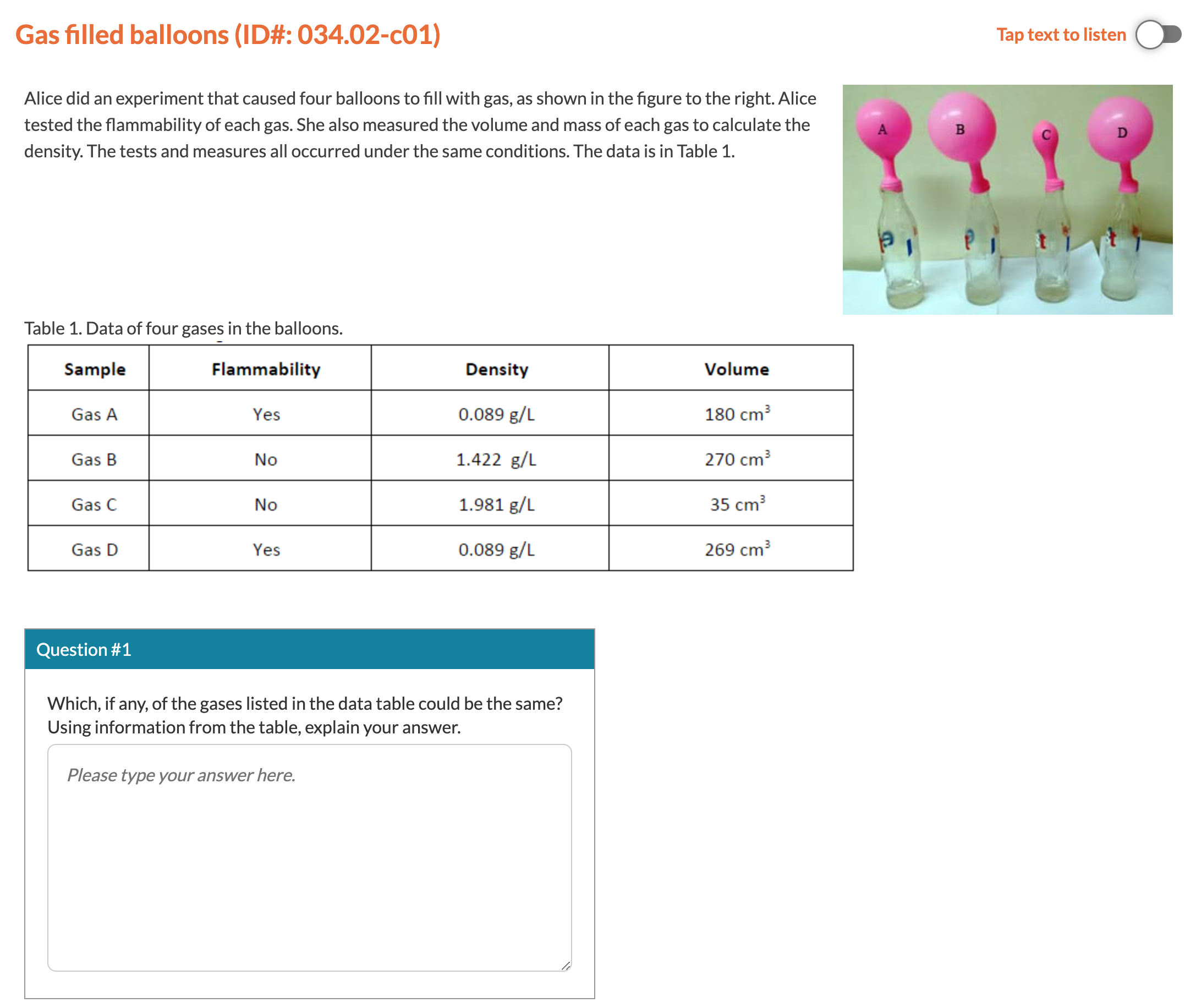}
    \caption{Illustrative Multi-label Task: Gas-Filled Balloons}
    \label{fig:gas_filled_ballons}
\end{figure}

The tasks draw on fundamental concepts of chemistry within the physical sciences domain, specifically “Chemical Reactions”. They require students to analyze data and distinguish substances by properties, capturing multi-dimensional reasoning in real-world contexts. Automated rubric-based scoring provides diagnostic information, highlighting areas where students may need additional support and offering educators actionable insights into student understanding.

\noindent\textbf{Task Example.}
For instance, in Task 3, students were asked to identify gases in an experiment by comparing their properties to those listed in a data table (see Fig. \ref{fig:gas_filled_ballons}). Successfully completing this task required understanding the structure and properties of matter, knowledge of chemical reactions, and the ability to plan investigations while recognizing patterns in the data.

A structured scoring rubric was developed to encompass five response dimensions aligned with the science learning framework: SEP+DCI, SEP+CCC-1, SEP+CCC-2, DCI-1, and DCI-2. The rubric was designed to capture multi-dimensional cognitive processes \cite{He2024G}. Table \ref{table:rubric_gas_filled_ballon} outlines the specific criteria for each dimension. Students were assessed simultaneously across these perspectives, receiving scores that reflected their understanding of DCIs, CCCs, and SEPs as defined by the rubric. To enhance the validity of these multi-perspective rubrics, the research team collaborated closely with experienced science educators.

\begin{table}[t]
\centering
\resizebox{\columnwidth}{!}{%
\begin{tabular}{l|l|p{0.66\columnwidth}}
\toprule
\textbf{ID} & \textbf{Perspective} & \textbf{Description} \\
\midrule
E1 & SEP+DCI & Student states that Gas A and D could be the same substance. \\
E2 & SEP+CCC-1 & Student describes the pattern (comparing data in different columns) in the flammability data of Gas A and Gas D as the same. \\
E3 & SEP+CCC-1 & Student describes the pattern (comparing data in different columns) in the density data of Gas A and Gas D, which is the same in the table. \\
E4 & DCI-2 & Student indicates flammability is one characteristic of identifying substances. \\
E5 & DCI-2 & Student indicates that density is one characteristic of identifying substances. \\
\bottomrule
\end{tabular}%
}
\caption{Scoring rubric for Task 3, Gas-filled balloons.}
\label{table:rubric_gas_filled_ballon}
\end{table}


\section{Experiments}\label{sec:exp}

We evaluate on seven NGSS-aligned, multi-label scoring tasks (Tasks 1–7), and reserve two additional tasks (Tasks 8–9) to test adaptation to new assessments (see Table~\ref{table:data} for label counts and split sizes).
Each task consists of short, open-ended science responses with rubric-aligned binary indicators (multi-label). We use the official train/test partitions and carve a stratified validation split (10\%) from training for model selection.
Unless noted, all experiments use BERT-base-uncased as the backbone encoder.

\noindent\textbf{Baselines} 
\begin{itemize}
    \item \textbf{Individual (Per-Task).} Separate BERT-base models fine-tuned per task (upper-bound capacity).
    \item \textbf{MTL (Shared Encoder + Heads).} A single BERT-base shared encoder with lightweight task-specific classification heads; no MoE.
    \item \textbf{UniMoE (Ours, no KD).} The MTL backbone is augmented with a gated Mixture-of-Experts (MoE) block placed between the encoder and heads.
    \item \textbf{UniMoE-Guided (Ours).} UniMoE trained with KD from task-specialized teachers.
\end{itemize}

\noindent\textbf{Evaluation Metrics} 
\begin{itemize}
    \item \textbf{Educational Reliability.} We follow educational scoring practice and report: Cohen's kappa score \citep{mchugh2012interrater} averaged over labels, Macro-F1, Micro-F1, and per-label accuracy averaged over labels. We use a fixed decision threshold of $0.5$ across all labels.
    \item \textbf{Resource Efficiency.} We report efficiency (parameter count, disk size) and latency (per-sample inference time).
    \item \textbf{Generalization Ability.} We evaluate the model's performance on held-out tasks (Tasks 8–9) using the educational reliability criterion introduced earlier.
\end{itemize}

\noindent\textbf{Model Structure.}
Our architecture builds on BERT-base-uncased as a shared encoder (109M, 91\%). We employ two MoE modules (9M, 7.3\%), with each containing a configurable number of experts (3–5 tested), with each expert designed as a two-layer feed-forward network. A gating mechanism with task-aware routing assigns inputs to experts, and load balancing is encouraged through a regularization penalty. Each task then uses a lightweight head, consisting of a linear projection, ReLU activation, dropout, and a final linear layer producing task-specific logits (2M, 1.7\% for all heads). KD integrates pre-computed teacher probabilities using temperature-scaled binary cross-entropy.


\begin{table*}[ht]
\centering
\resizebox{1.85\columnwidth}{!}{%
\begin{tabular}{l|l|rrrrrrrr}
\toprule
\textbf{Metric} & \textbf{Method} & \textbf{Task 1} & \textbf{Task 2} & \textbf{Task 3} & \textbf{Task 4} & \textbf{Task 5} & \textbf{Task 6} & \textbf{Task 7} & \textbf{Average} \\
\midrule

\multirow{4}{*}{\textbf{Cohen's $\kappa$ $\uparrow$}}
& \textcolor{blue}{Individual} & \textcolor{blue}{0.5164} & \textcolor{blue}{0.5526} & \textcolor{blue}{0.8638} & \textcolor{blue}{0.7140} & \textcolor{blue}{0.6863} & \textcolor{blue}{0.6480} & \textcolor{blue}{0.5740} & \textcolor{blue}{0.6510} \\
& MTL            & 0.4277 & 0.5052 & 0.8891 & 0.5238 & 0.5782 & 0.4393 & 0.5281 & 0.5559 \\
& UniMoE (no KD)    & 0.4457 & 0.5454 & \textbf{0.9091} & 0.6418 & 0.7340 & 0.6012 & \textbf{0.6063} & 0.6405 \\
& UniMoE-Guided   & \textbf{0.5246} & \textbf{0.5691} & 0.8830 & \textbf{0.6499} & \textbf{0.7545} & \textbf{0.6288} & 0.5787 & \textbf{0.6555} \\
\midrule

\multirow{4}{*}{\textbf{Macro-F1 $\uparrow$}}
& \textcolor{blue}{Individual} & \textcolor{blue}{0.5636} & \textcolor{blue}{0.6725} & \textcolor{blue}{0.9574} & \textcolor{blue}{0.7952} & \textcolor{blue}{0.7439} & \textcolor{blue}{0.8007} & \textcolor{blue}{0.6831} & \textcolor{blue}{0.7444} \\
& MTL            & 0.4832 & 0.6325 & 0.9662 & 0.6909 & 0.6441 & 0.6644 & 0.6415 & 0.6747 \\
& UniMoE (no KD)    & 0.4857 & 0.6659 & \textbf{0.9727} & 0.7541 & 0.7952 & 0.7697 & \textbf{0.7027} & 0.7351 \\
& UniMoE-Guided   & \textbf{0.5806} & \textbf{0.6958} & 0.9657 & \textbf{0.7603} & \textbf{0.8133} & \textbf{0.7939} & 0.6859 & \textbf{0.7565} \\
\midrule

\multirow{4}{*}{\textbf{Micro-F1 $\uparrow$}}
& \textcolor{blue}{Individual} & \textcolor{blue}{0.8859} & \textcolor{blue}{0.7725} & \textcolor{blue}{0.9577} & \textcolor{blue}{0.8746} & \textcolor{blue}{0.7891} & \textcolor{blue}{0.8551} & \textcolor{blue}{0.7469} & \textcolor{blue}{0.8290} \\
& MTL            & 0.8740 & 0.7860 & 0.9660 & 0.7813 & 0.7236 & 0.8096 & 0.7341 & 0.8107 \\
& UniMoE (no KD)    & \textbf{0.8911} & 0.7443 & 0.9620 & \textbf{0.8360} & 0.8089 & 0.8422 & \textbf{0.7669} & 0.8359 \\
& UniMoE-Guided   & 0.8727 & \textbf{0.7884} & \textbf{0.9649} & 0.8349 & \textbf{0.8214} & \textbf{0.8487} & 0.7490 & \textbf{0.8400} \\
\midrule

\multirow{4}{*}{\makecell{\textbf{Per-Label}\\\textbf{Accuracy $\uparrow$}}}
& \textcolor{blue}{Individual} & \textcolor{blue}{0.9383} & \textcolor{blue}{0.8563} & \textcolor{blue}{0.9431} & \textcolor{blue}{0.9100} & \textcolor{blue}{0.9038} & \textcolor{blue}{0.8750} & \textcolor{blue}{0.8452} & \textcolor{blue}{0.8923} \\
& MTL            & 0.9310 & 0.8533 & 0.9542 & 0.8379 & 0.8705 & 0.8303 & 0.8351 & 0.8732 \\
& UniMoE (no KD)    & \textbf{0.9425} & 0.8488 & 0.9525 & \textbf{0.8846} & 0.9095 & 0.8648 & \textbf{0.8611} & \textbf{0.8948} \\
& UniMoE-Guided   & 0.9289 & \textbf{0.8578} & \textbf{0.9528} & 0.8825 & \textbf{0.9143} & \textbf{0.8699} & 0.8469 & 0.8933 \\
\bottomrule
\end{tabular}
}
\caption{Results on Tasks 1–7 across four methods. Each column reports performance per task, with averages shown in the final column. Metrics include Cohen’s $\kappa$, Macro-F1, Micro-F1, and Per-Label Accuracy. Best results per task are bolded.}\label{tab:all_results}
\end{table*}

\noindent\textbf{Model Training.}
All methods share the same basic training setup.
We use AdamW ($\text{lr}=2 \times 10^{-5}$, batch size=$32$, max epochs=$20$) with linear learning rate warmup and decay, gradient clipping, and early stopping (patience=$3$).
Input text is tokenized with the standard BERT tokenizer, truncated or padded to a maximum length of 100 tokens. Labels are represented as multi-hot binary vectors, and teacher probabilities are normalized for KD. 
Beyond this common setup, UniMoE and \textsc{UniMoE-Guided} require additional hyperparameter tuning because they integrate knowledge from multiple datasets and rely on weighting mechanisms.
For UniMoE, we search over the number of experts $M$ ($\{3, 4, 5\}$) and load balance weight $\lambda_{\operatorname{LB}}$ ($\{0.005, 0.01, 0.05\}$). For KD-UniMoE, we additionally tune the KD weight $\lambda_{\operatorname{KD}}$ ($\{0.05, 0.1, 0.3, 0.5\}$).
Experiments are run on NVIDIA Tesla V100 GPUs.

\noindent\textbf{Teacher Models (20B).}
For each task, we fine-tune the 20B-parameter \texttt{openai/gpt-oss-20b} model using parameter-efficient LoRA adapters (rank $r=32$, $\alpha=64$), which update about $1\%$ of the model’s weights while keeping the rest frozen. 
Training is performed on an NVIDIA H100 GPU in mixed precision. The resulting teachers provide task-specific probability distributions, which are then distilled into \textsc{UniMoE-Guided} training.

\subsection{Results}

To provide insight that directly calling commercial LLM APIs is insufficient for our educational setting, we evaluate few-shot prompting of the 20B teacher model using the question, analytic rubric, and a concise exemplar response. As detailed in the Supplementary Materials, the resulting average Cohen’s $\kappa$ is only about 0.05, demonstrating that off-the-shelf LLM prompting performs poorly on our tasks and that fine-tuned, rubric-aware models remain essential.

\noindent\textbf{Educational reliability.} Table~\ref{tab:all_results} summarizes the results across all evaluation metrics and tasks. Several key trends emerge. 
First, the naive \textsc{UniMoE} consistently outperforms the standard MTL baseline. For example, Cohen’s $\kappa$ improves from $0.556$ to $0.641$, indicating that expert routing effectively mitigates negative transfer and enhances rubric fidelity. Similar gains are observed for Macro-F1 ($0.675$ vs.\ $0.735$) and Micro-F1 ($0.811$ vs.\ $0.836$), confirming that MoE provides additional capacity without inflating computational cost.
Second, incorporating knowledge distillation (KD) further improves model performance. The \textsc{UniMoE-Guided} variant achieves the highest average scores across $\kappa$, Macro-F1, and Micro-F1 metrics, while matching the naive \textsc{UniMoE} in per-label accuracy. Notably, it even surpasses individually fine-tuned per-task BERT models in some settings. For example, \textsc{UniMoE-Guided} attains an average $\kappa$ of $0.656$, exceeding the individual models’ $0.651$, and achieves the best Macro-F1 of $0.757$ compared to $0.744$ from individual models. These results demonstrate that distilling knowledge from large task-specific teachers into a single multi-task MoE student not only closes the performance gap but can also surpass individually trained models while offering superior efficiency in storage and deployment.

\begin{figure}[hbt!]
    \centering
    \includegraphics[width=0.88\linewidth]{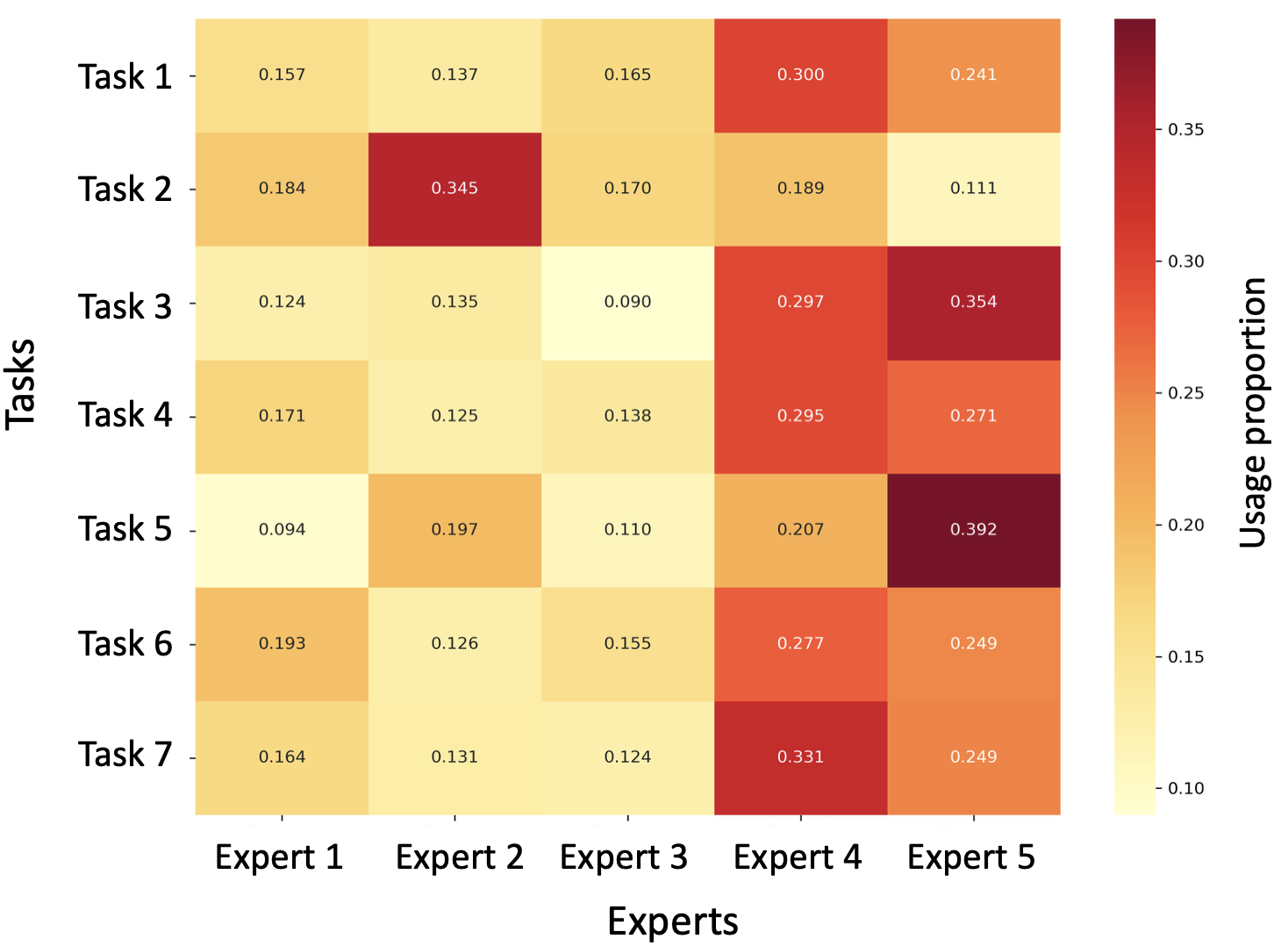}
    \caption{Expert usage patterns. Heatmap showing the proportion of each expert's usage for each task. }
    \label{fig:expert_usage}
\end{figure}

\begin{table*}[ht!]
\centering
\begin{tabular}{lcccc}
\toprule
\textbf{Model} & \textbf{Parameters} & \textbf{Disk Size (MB)} & \textbf{ Relative Size} & \textbf{Inference Time} \\
\midrule
UniMoE-Guided & 120M & 459  & 1.00x & 14 ms/sample \\
Teacher (oss)     & 20B & 40k (16-bit) & 87.15x & 43 ms/sample \\
Individual     & 772M & 2945 & 6.42x & 12 ms/sample \\
MTL          & 113M & 426  & 0.93x  & 13 ms/sample \\
\bottomrule
\end{tabular}
\caption{Model size and inference latency. Relative size is normalized to \textsc{UniMoE-Guided}. Latency is measured on a V100 GPU for all student models and on an H100 GPU for the teacher (not directly comparable).}\label{tab:efficiency}
\end{table*}

Figure~\ref{fig:expert_usage} shows the expert usage distribution across tasks in our \textsc{UniMoE-Guided} model. The results indicate that different tasks rely on experts with distinct preferences rather than uniformly distributing attention. For example, Task 2 heavily utilizes Expert 2 (34.5\%), while Task 5 shows a strong reliance on Expert 5 (39.2\%). Other tasks, such as Task 1 and Task 6, exhibit a more balanced distribution across multiple experts. This demonstrates that the MoE structure successfully enables task-specific specialization while still allowing shared experts to contribute across tasks, thereby capturing both commonalities and differences among diverse scoring rubrics.

\noindent\textbf{Resources Efficiency.}
Table~\ref{tab:efficiency} reports the efficiency and inference time of different model architectures. Our proposed \textsc{UniMoE-Guided} method achieves a compact size of 459 MB with 120M parameters, while maintaining an inference latency of 14 ms per sample. Compared to training individual BERT-based models for each task (772M parameters, 2945 MB), \textsc{UniMoE-Guided} reduces disk storage by more than 6$\times$. Similarly, relative to a simple MTL model (113M parameters, 426 MB), \textsc{UniMoE-Guided} shows nearly identical latency (14 ms vs.\ 13 ms), demonstrating that the additional MoE structure does not incur significant runtime cost. In contrast, the large GPT-based teacher model (20B parameters, 40 GB at FP16 on an H100 GPU) requires 87$\times$ more storage, highlighting the impracticality of direct deployment in the classroom or large-scale educational settings. While the inference time is reported, we load the model on an H100 due to its size, whereas student models run on a V100; therefore, the latency numbers are not directly comparable.
These results confirm that \textsc{UniMoE-Guided} strikes a favorable balance between efficiency and inference speed, while retaining the benefits of knowledge distilled from powerful teachers.

\subsection{Generalization}

\begin{table}[!b]
\centering
\caption{Generalization performance on held-out tasks (Task 8-9). Per-task models serve as an upper bound; MTL struggles to adapt, while \textsc{UniMoE-Guided} approaches per-task performance with only 7.5\% of parameters updated. }
\label{tab:generalization}
\begin{tabular}{l| l| c c}
\toprule
\textbf{Metric} & \textbf{Method} & \textbf{Task 8} & \textbf{Task 9} \\
\midrule
\multirow{3}{*}{Cohen’s $\kappa \uparrow$} 
  & Individual & 0.6724 & 0.6314 \\
  & MTL & 0.2954 & 0.2221 \\
  & UniMoE-Guided & 0.4680 & 0.4540 \\
\midrule
\multirow{3}{*}{Macro-F1 $\uparrow$} 
  & Individual & 0.7696 & 0.7130 \\
  & MTL & 0.5868 & 0.2707 \\
  & UniMoE-Guided & 0.6513 & 0.5592 \\
\midrule
\multirow{3}{*}{Micro-F1 $\uparrow$} 
  & Individual & 0.8595 & 0.7196 \\
  & MTL & 0.7077 & 0.2736 \\
  & UniMoE-Guided & 0.7543 & 0.5697 \\
\midrule
\multirow{3}{*}{\makecell{Per-Label\\Accuracy $\uparrow$}}
  & Individual & 0.8858 & 0.8729 \\
  & MTL & 0.7625 & 0.8153 \\
  & UniMoE-Guided & 0.8100 & 0.8297 \\
\bottomrule
\end{tabular}
\end{table}

We evaluate model generalization on two new tasks (Task 8 and Task 9), comparing against traditional MTL and individually trained models (gold standard). In MTL, a new head is added with the backbone frozen. In \textsc{UniMoE-Guided}, the backbone remains frozen while both a new head and MoE layers are fine-tuned, yielding $\sim$7.5$\%$ trainable parameters.
Results, as shown in Table \ref{tab:generalization}, demonstrate that \textsc{UniMoE-Guided} substantially outperforms MTL across all metrics and approaches the performance of individual models. For example, in Task 9, Cohen’s $\kappa$ improves from 0.2221 (MTL) to 0.4540, and Macro F1 from 0.2707 to 0.5592, more than doubling generalization quality. Fine-tuning the MoE layers induces slight decreases on old tasks (Cohen's $\kappa$: 0.6555 $\rightarrow$ 0.6279; Macro-F1: 0.7565 $\rightarrow$ 0.7322; Micro-F1: 0.8400 $\rightarrow$ 0.8258; Per-label acc: 0.8933 $\rightarrow$ 0.8870), yet performance remains consistently higher than MTL.
Overall, these results underscore the method’s efficiency and robustness, showing that \textsc{UniMoE-Guided} achieves strong generalization without suffering from the severe degradation typical of conventional MTL.

\section{Conclusion}\label{sec:conclu}

We presented \textsc{UniMoE-Guided}, a knowledge-distilled multi-task Mixture-of-Experts scorer designed for resource-constrained educational deployment. By pairing a shared encoder with weighted experts and small task heads - and distilling from task-specific teachers - the approach matches or exceeds per-task baselines at a fraction of the cost.
Evaluated across seven NGSS-aligned science tasks, \textsc{UniMoE-Guided} achieves comparable or superior performance to individually fine-tuned models (e.g., $\kappa$: 0.656 vs. 0.651; Macro-F1: 0.757 vs. 0.744), despite being $\sim$6$\times$ smaller than separate task-specific models and $\sim$87$\times$ smaller than the original 20B-parameter teacher model. This compact single-checkpoint design ensures practical classroom deployment, balancing speed, storage efficiency, and scoring accuracy.
A key advantage of \textsc{UniMoE-Guided} is its extensibility. For new assessments, simply adding a task head and lightly adapting the MoE (with the backbone frozen) yields significant improvements over MTL baselines—narrowing the gap to per-task specialization (e.g., held-out task performance: Cohen’s $\kappa$ 0.454 vs. 0.222 for MTL; Macro-F1 0.559 vs. 0.271). While minor regressions occur on prior tasks during adaptation, performance remains consistently above MTL levels, enabling low-effort incorporation of new rubrics without compromising existing capabilities.

\noindent\textbf{Limitations and future work.} We aim to (i) strengthen lifelong learning to further curb residual forgetting during task additions, (ii) improve interpretability of expert specialization and explore rubric-aware routing, and (iii) enhance the KD strategy to better distill knowledge from teachers.
Together, these directions aim to strengthen the reliability and reach of automated scoring systems that must scale across grades, standards, and classrooms while remaining efficient and deployable.




\newpage

\section{Acknowledgments}
This work was partially supported by the Institute of Education Sciences (IES) [R305C240010], the U.S. National Science Foundation (NSF) [DMS-2138854, DMS-1925066, DMS-1903226, DMS-2124493, DMS-2311297, DMS-2319279, DMS-2318809, 2101104], and the National Institutes of Health (NIH) [R01GM152814]. 
Any opinions, findings, conclusions, or recommendations expressed in this material are those of the authors and do not necessarily reflect the views of the IES, NSF, or NIH.

\bibliography{aaai2026}


\newpage

\onecolumn

\newpage
\begin{center}
\LARGE \textbf{Supplementary Material: Zero-Shot Prompting Baseline}\label{app:zeroshot}
\end{center}


\subsection{Setup}
We evaluate a frozen \texttt{openai/gpt-oss-20b} model in a zero-shot, multi-label setting on nine NGSS-aligned science tasks. For each dataset, we build an \emph{instruction preamble} by concatenating the task prompt, a concise exemplar response, and the analytic rubric. For every rubric element (label), we issue a separate binary query:

\begin{quote}
\textbf{You are an expert classifier. Answer strictly ``Yes'' or ``No''.}

\noindent\textbf{Text:} \(\langle\)student response\(\rangle\)

\noindent\textbf{Question:} Does the label ``\(\langle\)label name\(\rangle\)'' apply?

\noindent\textbf{Answer:}
\end{quote}

We compute \(P(\text{Yes})\) by summing next-token softmax mass over tokenizer IDs corresponding to \{``Yes'', ``yes'', ``YES'', ``1'', ``True'', ``true''\} and normalize by the mass over the union of \{Yes, No\} token sets. A fixed threshold of \(0.5\) converts probabilities to binary predictions. We report Micro-/Macro-F1, Exact Match at threshold \(0.5\) (denoted \emph{thr*}), Cohen’s \(\kappa\) averaged over labels, and mean Per-Label Accuracy.

\subsection{Rubric Integration Example (Task 3: Gas-Filled Balloons)}
This example mirrors the presentation in Section~\ref{sec:dataset} but shows precisely how we embed the scoring rubric into the zero-shot prompt.

\paragraph{Task prompt (abridged).}
Alice tested four gases (A--D), recording flammability, density, and volume under identical conditions. \emph{Question:} Which gases, if any, could be the same? Use the table to explain.

\paragraph{Exemplar rationale (abridged).}
Gases A and D could be the same because both are flammable and have the same density (\(0.089\,\text{g/L}\)); these are identifying properties.

\paragraph{Analytic rubric (embedded as plain text in the preamble).}
\setlength{\tabcolsep}{4pt}
\renewcommand{\arraystretch}{1.1}
\begin{table}[hbt!]
\centering
\caption{Scoring rubric for Task 3 (Gas-Filled Balloons), formatted to fit a single column.}
\label{tab:rubric_task3_appendix}
\begin{tabular}{l l p{0.60\columnwidth}}
\toprule
\textbf{ID} & \textbf{Perspective} & \textbf{Description} \\
\midrule
E1 & DCI & Student states that Gases A and D could be the same substance. \\
E2 & SEP+CCC & Supports the claim by pointing out that A and D have the same \emph{flammability}. \\
E3 & SEP+CCC & Supports the claim by pointing out that A and D have the same \emph{density}. \\
E4 & DCI & Indicates that \emph{flammability} is a characteristic property for identifying substances. \\
E5 & DCI & Indicates that \emph{density} is a characteristic property for identifying substances. \\
\bottomrule
\end{tabular}
\end{table}

\paragraph{Per-label question instantiation.}
Each rubric element becomes its own Yes/No query by substituting the label text:
\begin{quote}
\textbf{Question:} Does the label ``E2: A and D have the same flammability (pattern across columns)'' apply? \\ \textbf{Answer:}
\end{quote}
This produces one probability per label, later thresholded at \(0.5\).

\subsection{Results}
Table~\ref{tab:zs_results} summarizes zero-shot performance across all tasks. As expected, some items exhibit high mean per-label accuracy but low F1 (e.g., \emph{dry\_ice\_model}), reflecting label imbalance and conservative predictions. For \emph{gas\_filled\_balloons}, Cohen's \(\kappa\) is undefined (NaN) due to degenerate predictions on at least one label.


\begin{table}[hbt!]
\centering
\caption{Zero-shot results with \texttt{gpt-oss-20b} (frozen).}
\label{tab:zs_results}
\resizebox{0.7\columnwidth}{!}{%
\begin{tabular}{lcccc}
\toprule
\textbf{Dataset} &
\makecell{\textbf{Cohen's}\\\textbf{$\kappa$ @0.5}} &
\makecell{\textbf{Macro}\\\textbf{F1 @0.5}} &
\makecell{\textbf{Micro}\\\textbf{F1 @0.5}} &
\makecell{\textbf{Mean Per-}\\\textbf{label Acc @0.5}} \\
\midrule
anna\_vs\_carla                    & 0.1708 & 0.2261 & 0.1650 & 0.735375 \\
breaking\_down\_hydrogen\_peroxide & 0.1624 & 0.2850 & 0.2908 & 0.700600 \\
carlos\_javier\_atomic\_model      & 0.0561 & 0.1143 & 0.0951 & 0.603340 \\
dry\_ice\_model                    & 0.0000 & 0.0000 & 0.0000 & 0.792567 \\
gas\_filled\_balloons              & 0.0000 & 0.0000 & 0.0000 & 0.333333 \\
layers\_in\_test\_tube             & 0.0000 & 0.5007 & 0.5281 & 0.358760 \\
model\_for\_making\_water          & 0.0937 & 0.2080 & 0.2006 & 0.734280 \\
namis\_careful\_experiment         & 0.0000 & 0.0000 & 0.0000 & 0.566067 \\
natural\_sugar                     & 0.0119 & 0.2743 & 0.2783 & 0.605040 \\
\midrule
\textbf{Average}                  & \textbf{0.0543} & \textbf{0.1787} & \textbf{0.1731} & \textbf{0.6033} \\
\bottomrule
\end{tabular}%
}
\end{table}

\end{document}